\documentclass[journal]{IEEEtran}
\usepackage{graphicx}
\usepackage{comment}
\usepackage{amsmath,amssymb}
\usepackage{color}
\usepackage{multirow}
\usepackage{floatrow}
\newfloatcommand{capbtabbox}{table}[][\FBwidth]

\ifCLASSINFOpdf
\else
\fi

\begin{document}
%
\title{Synthetic Training for \\Monocular Human Mesh Recovery}
%
%
%
\markboth{ VOL. XX, NO. XX, XX 2020}{Yu \MakeLowercase{\textit{et al.}}: Synthetic Training for Monocular Human Mesh Recovery}

\author{ Yu Sun$^\star$,
Qian Bao,
 Wu Liu,
 Wenpeng Gao,
 Yili Fu,
 Chuang Gan,
 and~Tao Mei
\thanks{$^\star$This work was done when Yu Sun was an intern at JD AI Research.}}
 
 \maketitle

\begin{abstract}

Recovering 3D human mesh from monocular images is a popular topic in computer vision and has a wide range of applications. 
This paper aims to estimate 3D mesh of multiple body parts (e.g., body, hands) with large-scale differences from a single RGB image. 
Existing methods are mostly based on iterative optimization, which is very time-consuming. 
We propose to train a single-shot model to achieve this goal.
The main challenge is lacking training data that have complete 3D annotations of all body parts in 2D images.
To solve this problem, we design a multi-branch framework to disentangle the regression of different body properties, enabling us to separate each component's training in a synthetic training manner using unpaired data available. 
Besides, to strengthen the generalization ability, most existing methods have used in-the-wild 2D pose datasets to supervise the estimated 3D pose via 3D-to-2D projection. 
However, we observe that the commonly used weak-perspective model performs poorly in dealing with the external foreshortening effect of camera projection. 
Therefore, we propose a depth-to-scale (D2S) projection to incorporate the depth difference into the projection function to derive per-joint scale variants for more proper supervision. 
The proposed method outperforms previous methods on the CMU Panoptic Studio dataset according to the evaluation results and achieves comparable results on the Human3.6M body and STB hand benchmarks. More impressively, the performance in close shot images gets significantly improved using the proposed D2S projection for weak supervision, while maintains obvious superiority in computational efficiency. 
\end{abstract}

\IEEEpeerreviewmaketitle

\section{Introduction}~\label{sec:intro}

%
%
%
%
\IEEEPARstart{H}{uman} 3D mesh recovery (reconstruction/modeling) is a hot and challenging research topic in the computer vision community, which has drawn more and more attention recently. Different from the general pose estimation that detects several 2D or 3D keypoints, the 3D human mesh contains thousands of vertices that can provide subtle cues for understanding human posture, behavior, and interaction. It is essential for many applications such as virtual try-on, human behavior understanding, human-scene interpretation, etc.
The early works focused on integrating 3D body information collected from a multi-view or RGB-D camera system~\cite{de2008performance,vlasic2008articulated} which is expensive and not easily available. Recently, more works try to directly recover the 3D human body mesh from a single image~\cite{keep,hmr,gcmr,humanshape,sun2019dsd-satn} to facilitate application. Besides, a more challenging problem has targeted 3D face~\cite{flame,ringnet,zollhofer2018state} and hand~\cite{hand_graph,de2011model,smplh} recovery from monocular face/hand images. Undoubtedly, the next break-through will be the integration of single-body-part 3D mesh recovery tasks within a unified framework, which leads to recover multiple body parts from a single image. We think it is much closer to the actual application scenario. Obviously, due to the image blur, occlusion, and truncation in real scenes, it is much more challenging.

\begin{figure*}
\hsize=\textwidth
\centering
 \includegraphics[width=.96\textwidth]{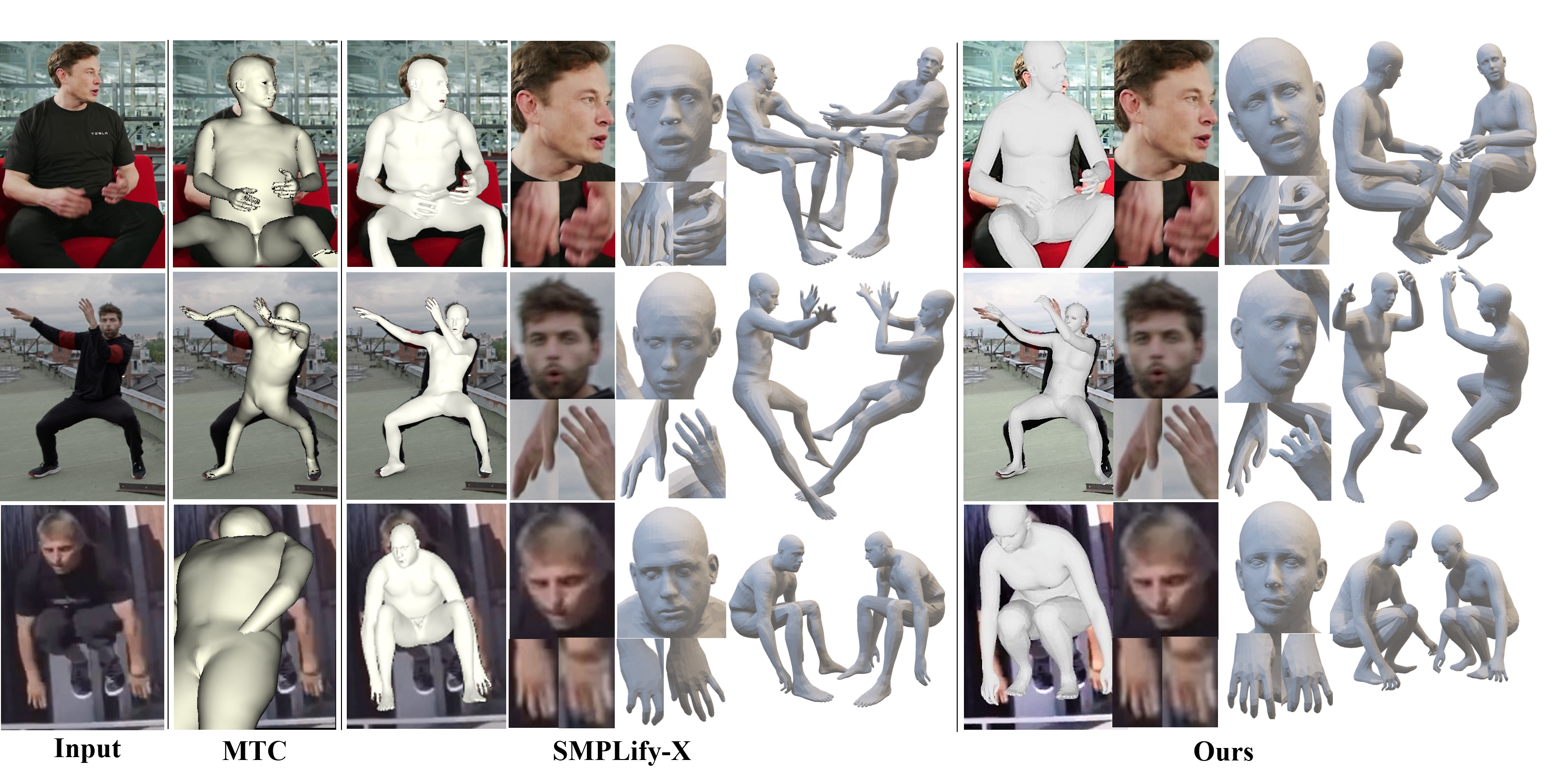}
\caption{In this paper, we present a synthetic training framework for monocular human mesh recovery using unpaired data. Compared with the previous methods, MTC~\cite{monocualar} and SMPLify-X~\cite{smplx}, model trained with proposed method shows obvious advantages in both performance and computational efficiency.}
\label{fig:demo1}
\end{figure*}

Previous works mainly focus on the single-body-part mesh recovery from monocular images, while research for the multi-part body mesh recovery is still at a relatively early stage. 
Existing works~\cite{monocualar,smplx} adopt a multi-stage optimization-based framework. They separately estimate the 2D/3D keypoints of each part using the individual module, and then make iterative optimizations on the statistical body model (e.g.SMPL-X~\cite{smplx}, Adam~\cite{totalcapture}) to fit the keypoint positions. They do not need much training data but rely on the multi-stage optimization strategy. Although the results seem promising, the iterative optimization process is very time-consuming. It takes tens of seconds to process an image, which makes it difficult to meet actual needs. Therefore, in this paper, we explore a neural network to reduce the run-time to milliseconds and keep the accuracy stand.

\begin{figure*}[t]
    \centering
    \includegraphics[width=0.98\columnwidth]{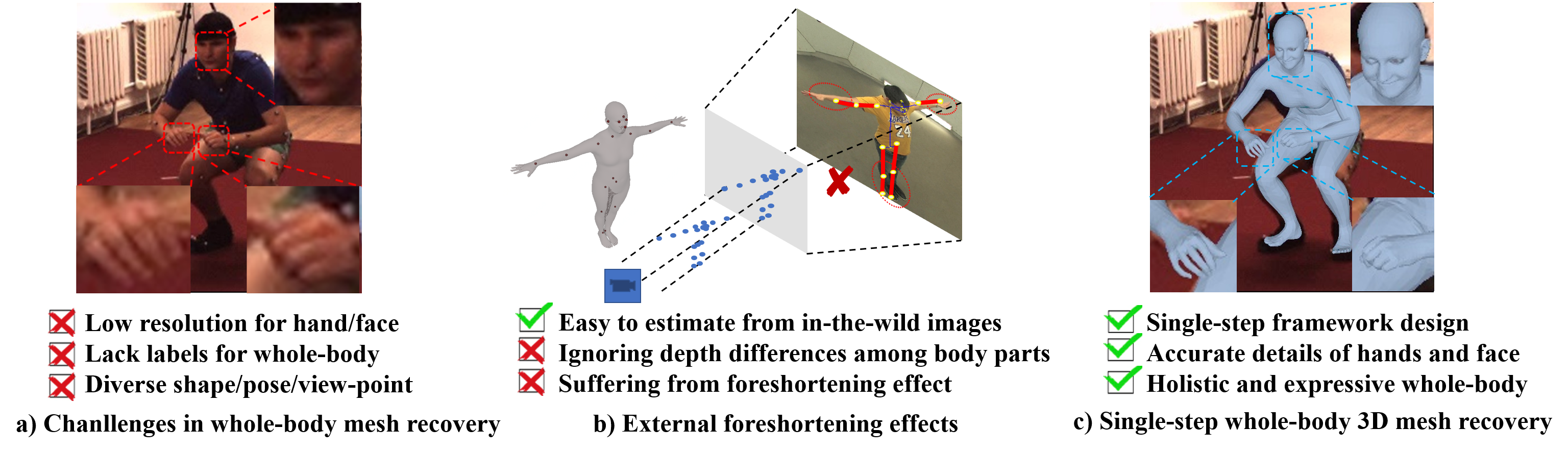} 
    \caption{Compared with separate 3D mesh recovery, the whole-body 3D mesh recovery faces the challenges of 1) the inherent physical size diversity shown in (a), and 2) the external foreshortening effect of camera projection in (b). The proposed framework can recover the holistic and expressive whole-body 3D mesh with the single-step framework in nearly real-time speed.}
    \label{Figure1}
\end{figure*}

The challenges of applying a learning-based algorithm on this task mainly come from two aspects, data lacking and large scale differences, as shown in Fig.~\ref{Figure1} (a). Firstly, we lack data with paired 2D images and 3D annotations for training. Due to the inherent physical size difference, it is difficult to precisely capture large torsos and small body parts at the same time. Therefore, there are merely datasets that contain 2D images with whole-body 3D mesh and motion annotations. 
In contrast, there are some single-body-part datasets of 3D body pose, hands gesture, and face expression available, which contain rich annotations of the individual parts in constrained experimental environments. In addition, due to the specific capturing configurations, sparse shape/pose space, large scale difference, et al., it is hard to directly combine them for training. To meet the need for data, we need to design a suitable framework that can be trained using the existing un-paired data.
Secondly, how to deal with the large scale differences between different human parts in one network is challenging.
Generally, we tend to recover the 3D hands/face/body from cropped high-resolution images with a centered target and most single-body-part 3D datasets~\cite{stb,Freihand2019,FECdataset,D3DFACS_dataset} are collected in this manner. However, as shown in Fig.~\ref{Figure1}, in general cases, the image patches of small body parts are likely to be blurred, partially invisible, and randomly placed in the whole-body image.
 The scale issue requires the network to handle different scale characteristics of each part. When recovering the whole-body 3D mesh, the body requires a large respective filed to learn the structure constraints, while the small parts, like hands, require a higher quality of input for complex pose prediction.

To address these problems, we design a multi-branch framework that can be trained using unpaired data in a synthetic training manner. Firstly, to deal with the large-scale differences between human parts, the network needs to provide a diverse receptive field for different body parts.  The key idea is to disentangle the estimation of different body properties, such as body pose, shape, and hand pose, in the convenience of separate training. Therefore, we design the framework with two branches that can learn the global and partial representation respectively. The global branch estimates the body shape and camera parameters with the high-level image features, while the partial branch achieves the estimation for small body parts, like 3D hand pose, individually. Benefited from this decoupled architecture design, each individual human part along with the holistic information can be recovered. 

Secondly, for the multi-branch framework training, we propose a synthetic training scheme (STS) to generate the paired data. Specifically, for the pose-related estimation in both branches, we employ the whole-body 2D pose predicted from the 2D image as an intermediate representation. In this way, the model gets trained to learn the mapping from 2D pose coordinates to 3D joint rotations. The training sample is supposed to contain a whole-body 2D pose along with the 3D rotation of each body joint. Since most existing datasets only contain 2D/3D pose of individual body parts that are scale different and captured in diverse configurations and shape/pose space, directly combing those individual data will cause weird poses. Instead, we propose STS to integrate them naturally and train the model via exploiting the consistency between the synthetic mesh and the regressed mesh. In this way, the STS helps to solve the lack of paired data for the directly supervised training, and meanwhile increases the diversity of the training data (e.g. full shooting angles, abundant shape/pose space, unified annotations) by combining datasets from multiple resources and generating data with controllable settings.

On the other hand, using synthetic data only for training brings the generalization problem. 
For instance, during training, we use synthetic data as input which are accurate and complete, while during inference the inputs are estimated 2D poses from in-the-wild images which are usually inaccurate and incomplete. 
The gap between synthetic and real may lead to poor in-the-wild performance. However, in-the-wild 3D datasets are hard to obtain.
To solve them, existing methods~\cite{hmr,gcmr,sun2019dsd-satn,ringnet,monocualar} usually employ in-the-wild 2D pose datasets for better generalization. In this process, due to the in-the-wild 2D images that are un-calibrated, weak-perspective camera model is widely adopted in the 3D-to-2D projection for  supervision. Nonetheless, in the weak-perspective model, the depth difference between different body parts is ignored.
As shown in Fig.~\ref{Figure1} (b), it may cause the failure of dealing with the foreshortening effects and makes the 3D-to-2D projection prone to misalignment. To cope with this problem, we design a D2S (D2S) projection function to calibrate the projection scale of each keypoint separately by involving the depth difference in the projection formulation. It can be regarded as a per-joint version of the weak-perspective model that can deal with the foreshortening effects and rectify the 2D pose supervision. 

The proposed method is well evaluated both qualitatively and quantitatively. As shown in Fig.~\ref{fig:demo1}, compared with previous SOTA methods, MTC~\cite{monocualar} and  SMPLify-X~\cite{smplx} , the model trained with the proposed synthetic training method shows an obvious advance in the estimation of body pose, shape, and expression. Especially benefited from getting rid of the iterative optimization process, the computational efficiency of our proposed single-step framework gets greatly improved. Our network requires only one forward pass to obtain results, while previous methods~\cite{monocualar,smplx}  need multiple iterative processes to optimize the complex expectation functions. 
We conduct experiments and evaluations on three benchmarks. State-of-the-art results are achieved on CMU Panoptic Studio dataset~\cite{monocualar}. In addition, compared with single part 3D recovery, comparative results are achieved on Human3.6M~\cite{h36m} body and STB hand dataset~\cite{stb}. Visualization results of the recovered expressive 3D human meshes also demonstrate the effectiveness of the proposed framework.

The main contributions of our paper are listed as follows:
\begin{itemize}
\item We design a decoupled framework for monocular human mesh recovery, which can be trained using unpaired data in a synthetic training manner.
\item A novel D2S projection is designed to deal with the camera foreshortening effects that the weak-perspective camera ignores in un-calibrated conditions.
\item Promising results are achieved on three relative benchmarks. The single-shot design shows obvious advantages in computational efficiency over the multi-stage iterative optimization-based methods~\cite{monocualar,smplx}.
\end{itemize}

\section{Related Work}\label{related work}
\textbf{Human Pose Estimation.} With the development of deep learning, a great improvement has been made in estimating keypoints of human skeleton, hands or face. 2D human pose estimation~\cite{openpose,whole-2d-openpose,TIP_2DPEnie2018hierarchical,TIP_2DPEluo2018multi,TIP_2DPEfu2016orgm} aims to detect 2D keypoint locations, while 3D human pose estimation~\cite{chen20173d,fang2018learning,nie2017monocular,soft-argmax,sun2017compositional,zhou2017towards,TIP_3DPEzheng2020joint} further infers the 3D locations from 2D image. 3D human pose estimation has shown the advantage of human spatial representation and there are two kinds of 3D human pose estimation approaches. One is the two-stage scheme that first detects 2D locations then infers 3D locations from the 2D keypoints input~\cite{chen20173d,fang2018learning,nie2017monocular}.
The other scheme directly predicts 3D human pose from images~\cite{soft-argmax,sun2017compositional,zhou2017towards}.  Human pose estimation can provide keypoint locations for skeleton description, however, the sparse location representation is not enough for understanding humans sufficiently. In our work, we use the 2D/3D keypoints descriptions as the intermediate supervision for our 3D mesh recovery framework. 

\textbf{3D Human Mesh Recovery. }
To reduce the complexity, many existing approaches regard this problem as recovering parameters of a statistical 3D human model. The 3D body statistical model SMPL~\cite{smpl} has been widely used because of its good performance and free access. Recently, plenty of ConvNet-based parameters recovery methods~\cite{hmr,humanshape,yoshiyasu2018skeleton,monocap,monoperfcap} are proposed. Compared with the optimization-based methods (e.g. SMPLify~\cite{keep}), ConvNet-based methods have shown obvious advantages in both performance and computational efficiency. 
However, these learning-based methods are limited by lacking paired data for training. Therefore, many methods focus on developing various loss functions for  supervising the model with data available.
For example, HMR~\cite{hmr} trained a convolution neural network using unpaired data in a generative adversarial manner. They employ a discriminator to distinguish the rationality of the predicted SMPL parameters using 3D motion capture data. 
DenseRaC~\cite{Xu_2019_ICCV} rendered the estimated 3D mesh (textured with the part segmentation map) back to the 2D plane and utilized the DensePose results for supervision.
	TexturePose~\cite{pavlakos2019texturepose} utilized the appearance consistency of the person among multi-viewpoints (or even adjacent video frames) for supervision.
    Kundu et. al.~\cite{Kundu_Appearance_Consensus} also employed the appearance consistency to make up the lack of paired training data but in a self-supervised manner. They have used  the image pairs of the same person to disentangle the foreground human appearance from the background and map it to body mesh for supervising the consistency of vertex-level appearance.
Besides, other methods seek to employ disentangled representation for intermediate supervision. For instance, Sun et al.~\cite{sun2019dsd-satn} takes the 2D pose as an intermediate representation and develop a skeleton-disentangled representation to tackle the feature coupling problem and reduce the task complexity. 
More and more researchers realize the importance of 2D poses for better generalization. 
Choi et. al.~\cite{Choi_Pose2Mesh} proposed a graph neural network to hierarchically regress the 3D human mesh from the 2D pose in a coarse-to-fine manner.
In this work, we employ 2D pose as an intermediate representation to connect the 3D space and 2D image plane.

\textbf{3D Hands and Face Mesh Recovery.} Modeling hands and faces are independent research areas from human body recovery. Statistical models such as MANO~\cite{smplh} and FLAME~\cite{flame} have been widely adopted to reduce complexity. Lacking a 3D dataset is still the main challenge.
Kulon et al.~\cite{Kulon_2020_CVPR} proposed to collect internet images with 3D hand mesh annotations via an iterative fitting hand model based on its 2D pose.
Zhou et al.~\cite{zhou2020monocular} developed a multi-stage framework for 3D hand mesh recovery, which can make use of all the sources of available training data, such as 2D/3D pose and 3D motion capture data. Especially, a model is trained to estimate MANO pose and shape parameters from the 3D pose position.
Besides, face modeling is promoted by 3D scans models \cite{zollhofer2018state}, which can represent various face shapes and expressions. FLAME \cite{flame} first models the whole head region instead of the face region solely, and outputs the head rotations along with the neck region. Based on FLAME, RingNet \cite{ringnet} designs an end-to-end 3D face and expression estimation network with a shape consistency loss to learn 2D-to-3D mapping. 
However, most existing methods for 3D hand/face mesh recovery only focus on estimating from the 2D  high-resolution cropped images of face/hand, which is different from the blurred whole-body image we used. 

\textbf{3D Whole-body Recovery with Body, Hands, and Face.} 
Different from the separate models, Joo et al.~\cite{totalcapture} introduced the Adam model to capture human motion by stitching the body, hands and, face model together.
SMPL+H \cite{smplh} integrates a 3D hand model into the SMPL body model while it does not consider face modeling. 
Xiang et al.~\cite{monocualar} use separate CNN networks for the body, hand, and face, and then jointly fits the Adam model to the outputs of all body parts using an optimization-based algorithm. 
A part orientation field~\cite{openpose,whole-2d-openpose} is adopted to encode 3D orientation of body parts w.r.t. 2D space. 
SMPL-X \cite{smplx} adds the FLAME head model \cite{flame} to SMPL+H and learns the shape and pose-dependent blend shapes by fitting the model to 3D scans data. Furthermore, SMPLify-X~\cite{smplx} is proposed to recover the full human 3D mesh by iterative fitting SMPL-X model to 2D keypoints of face, hands, and body. 
Those previous works promote the 3D body recovery with hands and faces. However, they are all based on the iterative optimization algorithm, which is time-consuming and not optimal. In this paper, our work further explores the single-step model for the body, hands, and face recovery. 

\section{Method}

\begin{figure*}[t]
    \centering
    \includegraphics[width=1\textwidth]{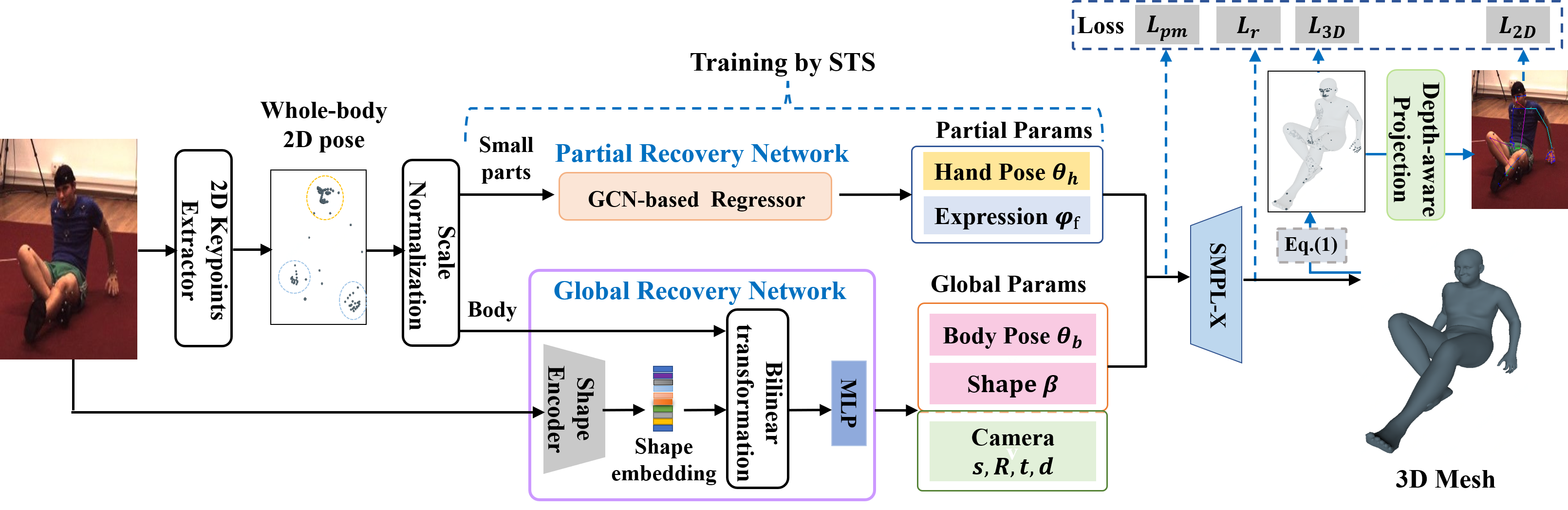}
    \caption{Overview of the proposed single-step whole-body 3D mesh recovery framework. The network contains two branches for retrieving the global (body and camera) and partial (face and hands) parameters separately. The supervision is carried out on the estimated parameters and 2D/3D keypoints of 3D mesh. D2S projection is developed to make a 3D-to-2D projection for accurate weak supervision. We employ the STS to train the GCN-based regressor.
    }
    \label{framework}
\end{figure*}
\subsection{Overview}
We aim to train a single-shot model for whole-body 3D human mesh recovery, including body, hands, and expression from a single uncalibrated image. By involving the statistical model SMPL-X~\cite{smplx} to encode the whole-body 3D human mesh, the task is converted to estimate the SMPL-X parameters from a 2D image.  Fig.~\ref{framework} shows an overview of the proposed framework. Take a single 2D image as input, the outputs of our network are global body/camera and partial hands/expression parameters for the SMPL-X model that derives the 3D mesh. 
In our proposed framework, a two-branch design is adopted to estimate parameters of 1) large-scale body and camera, 2) small-scale 3D hands pose, and facial expression, respectively. 
Since we lack the paired data to support the directly supervised training of the partial branch, we propose a synthetic training scheme (STS) to help regress hands and face parameters with the unpaired 3D data in a self-supervised manner. Besides, for better generalization and richer representation learning, we supervise the 3D pose of the estimated 3D mesh with 2D pose using the proposed D2S projection. We will introduce the network architecture and main modules of our framework in the following.

\subsection{Background\label{sec:background}}

\subsubsection{SMPL-X Model}

We employ a unified parametric model, SMPL-X~\cite{smplx}, for recovering the 3D human mesh with our estimated body, hands, and expression parameters as input.  In this part, we summarize the salient aspects of the model in our notation. SMPL-X can be regarded as the integration of the SMPL body model, FLAME head model, and MANO hand model~\cite{smplh}. 
In SMPL-X, statistical parameters of pose $\boldsymbol{\theta}$, shape $\boldsymbol{\beta}$, and expression $\boldsymbol{\psi}$ are disentangled and used to control the human 3D mesh variations of different aspects. An efficient mapping $M( \boldsymbol{\theta}, \boldsymbol{\beta}, \boldsymbol{\psi} ;\Phi):\mathbb{R}^{| \boldsymbol{\theta}|\times| \boldsymbol{\beta}|\times| \boldsymbol{\psi}|}\mapsto\mathbb{R}^{3 \times N}$ is established to recover the 3D human mesh with $N = 10, 475$ vertices, where $\Phi$ represents the statistical prior of human body mesh. 

The pose parameter $ \boldsymbol{\theta} \in \mathbb{R}^{3 \times (K+1)}$ represents the relative 3D rotation of $K=54$  keypoints, including poses of body $ \boldsymbol{\theta_b} \in \mathbb{R}^{3 \times 21}$, each hand $ \boldsymbol{\theta_h} \in \mathbb{R}^{3 \times 15}$, jaw $ \boldsymbol{\theta_j} \in \mathbb{R}^{3 \times 1}$, each eye $ \boldsymbol{\theta_e} \in \mathbb{R}^{3 \times 1}$ and global rotation $ \boldsymbol{\theta_{gr}} \in \mathbb{R}^{3 \times 1}$. The shape parameter $ \boldsymbol{\beta} \in \mathbb{R}^{10}$ represents the joint shape of body, face and hands. The expression parameter $ \boldsymbol{\psi} \in \mathbb{R}^{10}$ represents the facial expression. Both $\boldsymbol{\beta}$ and $\boldsymbol{\psi}$ are PCA coefficients that denote the variations of first 10 principle components. Here, for simplification, we use a single vector $\boldsymbol{\psi_f}  \in \mathbb{R}^{13}$ to represent the combination of $\boldsymbol{\psi}$ and $ \boldsymbol{\theta_j}$. $ \boldsymbol{\theta_{hp}} \in \mathbb{R}^{6}$  is a PCA coefficient for each hand to capture the pose variations of 15 keypoints (45 parameters). Besides, a linear regressor $P_{3D}$ is developed to derive the 3D keypoint locations $J_{3D} \in \mathbb{R}^{3 \times K}$ from vertices of human 3D mesh by
\begin{equation}
\begin{aligned}
J_{3D} = M( \boldsymbol{\theta}, \boldsymbol{\beta}, \boldsymbol{\psi} ;\Phi)P_{3D}.
\label{eq:j3d}
\end{aligned}
\end{equation}

The SMPL-X model encodes the complex human mesh into a one-dimensional parameter vector and decouples different body part factors (e.g. shape, pose, facial expression) into individual parameters. We use the mapping $M( \boldsymbol{\theta}, \boldsymbol{\beta}, \boldsymbol{\psi} ;\Phi)$ to recover the total 3D human mesh and those statistical parameters $ \boldsymbol{\theta}, \boldsymbol{\beta}, \boldsymbol{\psi} $ are yielded by our proposed network. 

\subsubsection{Perspective projection}\label{project function}

In general, a perspective model is employed to carry out the 3D-to-2D projection, establishing the mapping from real 3D space to 2D image. The projection from the $i$-th human 3D keypoint $J_i = (x_i,y_i,z_i)$  to image coordinates ${J^{p2d}_i } = (u_i,v_i)$ can be derived as 
\begin{equation}
\begin{aligned}
{J^{p2d}_i} = \boldsymbol{A} (\boldsymbol{R^c} {J_i} + {t^c}),
\end{aligned}
\label{eq:persp0}
\end{equation}
where $ \boldsymbol{A}$ is the camera intrinsic matrix,  $\boldsymbol{R^c}$ and $ \boldsymbol{t^c}$ are extrinsic parameters denoting the camera 3D rotation and translation respectively. For simplification, we integrate the camera rotation expression $\boldsymbol{R^c}$ into the human body rotation parameter $\boldsymbol{R}$ and ignore the camera distortion. In this way,  the coordinates of $J^{p2d}_i$ can be derived as
\begin{equation}
\begin{aligned}
u_i = \frac{f_x (x_i + t^c_x)}{d + z_i} = \frac{f_x}{d  + z_i} x_i  + \frac{f_x t^c_x}{d + z_i},
\end{aligned}
\label{eq:persp1}
\end{equation}
\begin{equation}
\begin{aligned}
v_i = \frac{f_y (y_i + t^c_y)}{d + z_i} = \frac{f_y}{d  + z_i} y_i  + \frac{f_y t^c_y}{d + z_i},
\end{aligned}
\label{eq:persp2}
\end{equation}
where $t^c = (t^c_x, t^c_y, d)$ is the 3D translation from body center to camera. Besides, $f_x = f/ dx$  and $f_y = f / dy$ where $f$ is the camera focal length, $dx$ and $dy$ are per-pixel physical sizes in the height and width dimension, respectively.

\subsubsection{Weak-perspective projection\label{weak-persp function}}

For the convenience of supervising 3D human mesh with 2D pose labels, it is  necessary to represent a 3D human body in camera coordinates. However, the input 2D images are un-calibrated and the complete perspective camera parameters are hard to retrieve. In this situation, weak-perspective camera model is widely used in most existing methods for calculating the 2D projection $J^{wp2d}_i$ of 3D keypoints $J_i$ by\begin{equation}
\begin{aligned}
J^{p2D}_{i} = s \rm{\Pi}(\boldsymbol{R}{J_i}) + \boldsymbol{t},
\end{aligned}
\label{eq:wpersp1}
\end{equation}
where $\boldsymbol{R} \in \mathbb{R}^{3}$ is the global rotation parameter, $\rm{\Pi}$ is an orthographic projection operation, $ \boldsymbol{t} \in \mathbb{R}^{2}$ and $ s\in \mathbb{R}$ represent translation and scale on the image plane, respectively. In detail, $J^{p2d}_i$ can be derived as 
\begin{equation}
\begin{aligned}
u_i = s x_{i} + t_u,
\end{aligned}
\label{eq:wpersp2}
\end{equation}
\begin{equation}
\begin{aligned}
v_i = s y_{i} + t_v,
\end{aligned}
\label{eq:wpersp3}
\end{equation}

\begin{figure*}
\hsize=\textwidth
\centering
 \includegraphics[width=.96\textwidth]{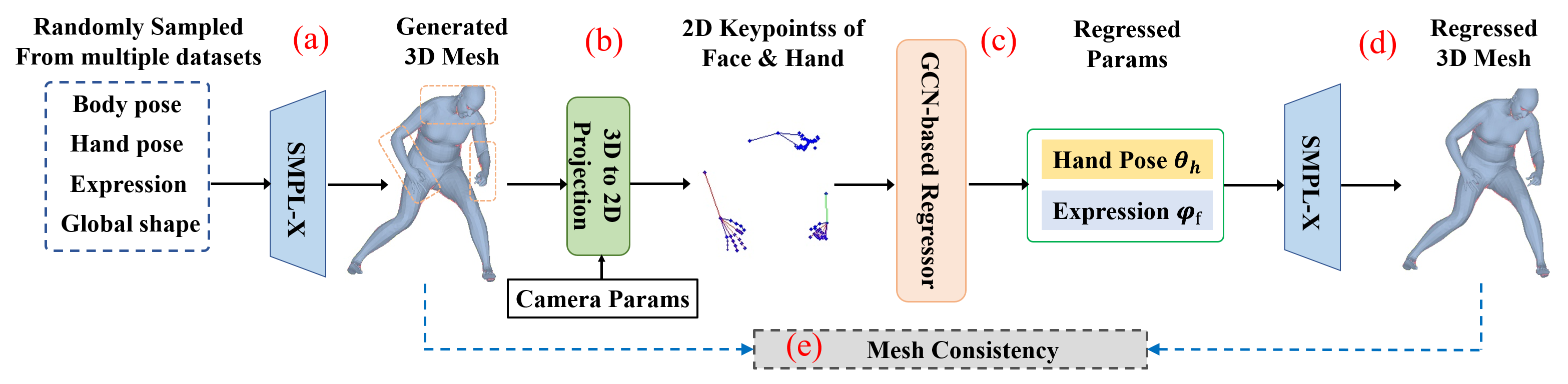} 
    \caption{The pipeline of the synthetic training scheme. Without full annotations, the face/hand partial recovery sub-network can be trained by supervising the consistency between the first generated 3D mesh and final regressed 3D mesh.}
    \label{self-super}
\end{figure*}

\subsection{3D Human Parameters Recovery Network}

As shown in Fig.~\ref{framework}, a two-branch framework is designed to predict the SMPL-X and camera parameters from a 2D image for whole-body 3D mesh recovery. Given 2D images, we use OpenPose~\cite{openpose} and a shape encoder to predict the whole-body 2D keypoints and shape embedding, respectively. Then, to balance the scale differences between different body parts, we develop a scale normalization layer to transform the 2D keypoints of the face, hands, and body into a standardized space. In detail, keypoints of different parts are first translated to their center by subtracting the mean value and then scaled by dividing the half size of their bounding boxes. Then the re-scaled  2D keypoints of body and hands/face are separately put into the two sub-networks for global and partial parameter retrieval.  

In global branch, with the re-scaled 2D body keypoints and the extracted shape embedding (i.e, 512-D global feature vector from Resnet-50~\cite{resnet}), we use the global recovery sub-network to estimate global parameters. In this process, following \cite{sun2019dsd-satn}, we employ a bilinear transformation layer to re-organize the global features for better performance. Then, we put the yielded skeleton-disentangled representation into a Multi-Layer Perceptron (MLP) to regress the body shape $\boldsymbol{\beta}$,  pose parameters $ \boldsymbol{\theta_b}$, and the camera parameters $(s,\boldsymbol{R},\boldsymbol{t}, d)$.  $ \boldsymbol{R} \in \mathbb{R}^{3}$ is the 3D rotation of the entire human mesh,   $ \boldsymbol{t} \in \mathbb{R}^{2}$ and $ s \in \mathbb{R}$ are translation and scale in image plane, respectively. $d \in \mathbb{R}$ is a distance factor of D2S projection, which will be introduced in Sec.\ref{sec:D2S proj}.

In partial branch, with the re-scaled 2D face and hand keypoints, we use the partial recovery sub-network to predict the small-scale 3D hands pose $\boldsymbol{\theta_h}$ and facial expression $\boldsymbol{\psi_f}$ parameters. Especially, since both face and hand has the structure prior (e.g., symmetry, connection), we leverage the GCNs to explore their structural and spatial relations. The design of the GCN-based face regressor follows~\cite{zhao2019semantic} and consists of four semantic graph convolution layers.  
Similarly,  the GCN-based hand regressor estimates hands pose parameters $\boldsymbol{\theta_h}$. The only difference is that the face regressor employs an additional MLP to estimate the expression from whole-face features.

With these estimated parameters from both the global and partial branches, we directly connect all these parameters and put into the SMPL-X model described in Sec.\ref{sec:background} to obtain the final 3D human mesh. In other word, the parameter vector $\mathbf{\Theta} = \{ \boldsymbol{\theta}, \boldsymbol{\beta}, \boldsymbol{\psi},  s,\boldsymbol{R},\boldsymbol{t},d\}$ is adopted to represent the 3D human whole-body in camera coordinates.

\textbf{Loss function.} The supervision of the proposed framework is conducted in both 3D and 2D to make full use of existing data. The total loss $L_{\mathbf{\Theta}}$ is the weighted sum of parameter $L_{pm}$, 3D keypoints $L_{3D}$, 2D projected keypoints $L_{2D}$ and the rationality $L_{r}$ where $w_{pm}, w_{3D}, w_{2D}, w_{r}$ are weights of these loss items.  
Since the ground truth parameters are not fully available in most datasets, the supervision is usually carried out on the different aspects of the recovered 3D human mesh. As stated in Eq.\ref{eq:j3d}, 3D keypoints can be derived from human mesh via a linear keypoint regressor. is employed to supervise the 3D keypoints of the estimated whole-body mesh using the  $L2$ loss. 

In 3D level, we employ $L2$ loss $L_{pm}$ and $L_{3D}$ to supervise the estimated parameters and the 3D poses (derived by Eq.\ref{eq:j3d}) of all human parts respectively. In detail, $L_{pm}$ is  employed to supervise the pose $\boldsymbol{\theta}$ and expression $\boldsymbol{\psi}$. The angular pose is converted from axis-angle representation to rotation matrices via the Rodrigues formula for more stable training~\cite{hmr,humanshape}. 
However, due to the inherent scale difference, it is hard to simultaneously capture large torsos and small body parts. Therefore, no dataset has full whole-body 3D annotations. In this case, general supervised training is not feasible. In contrast, there are some single-body-part datasets of 3D body pose, hand gesture, and facial expression available, which are not paired. Therefore, we develop a self-supervised learning scheme (STS) in Sec.~\ref{STS} to train the partial branch in a learning-from-synthetic manner. 

The existed datasets are mostly captured in restricted environments. The model solely trained on them is hard to work well in the wild. To solve this generalization problem, in the 2D level, we project the 3D joints of the estimated 3D mesh onto the 2D image plane using the estimated camera parameters for better generalization. The $L1$ loss $L_{2D}$ is used to make 2D pose supervision.
However, the widely used weak-perspective camera model ignores the foreshortening effects in 2D images, which makes it prone to fail when the images are captured closely. Therefore, we propose a novel D2S projection in Sec.\ref{sec:D2S proj} to make the projection scale adjustment for each keypoint in terms of the depth difference between them. Finally, following~\cite{hmr}, we also supervise the rationality $L_{r}$ of the keypoint angles and whole-body shape for preventing physically impossible 3D human mesh, especially when we use the unpaired body, hands, and face data.

\subsection{STS: synthetic training Scheme\label{STS}}
As we stated in Sec.~\ref{sec:intro} and Sec. 3.3, lacking complete annotations of whole-body parts makes it hard to directly train the network end-to-end. Therefore, we develop a synthetic training scheme (STS) to supervise the network under this circumstance. The STS is designed to get reasonable synthetic data from the original unpaired 3D data and learn the hands and face recovery in a self-supervised manner. 

The illustration of STS is shown in Fig.~\ref{self-super}. Firstly, we collect the 3D pose $\boldsymbol{\theta}$, shape $\boldsymbol{\beta}$ and expression $\boldsymbol{\psi_f}$ parameters from these unpaired datasets by fitting the statistical models to establish a reasonable parameter bank.
Specifically, we get facial expression $\boldsymbol{\psi_f}$ by fitting FLAME~\cite{flame} model to 3D head mesh data~\cite{D3DFACS_dataset} or 2D facial keypoints~\cite{bulat2017far}. The body poses are brought from several Mocap datasets~\cite{amass}, and 3D hands poses are brought from MANO datasets~\cite{smplh,Freihand2019,amass}. 
Then, as shown in Fig.~\ref{self-super} (a) we randomly sample these parameters from the bank to form the complete parameter vector $\boldsymbol{\Theta}$ and put it into the SMPL-X model to derive the whole-body 3D mesh. 
Next (b), we map the joints of the driven 3D whole body mesh back to the 2D image plane to establish the mapping between 3D mesh and 2D pose. In this way, 2D pose along with its complete 3D annotations are generated. 
With these synthetic paired data, in (c) and (d), we exploit the GCN-based regression network for estimating hand gesture and facial expression respectively from their 2D landmarks. Finally in (e), the face/hand partial recovery sub-network can be trained by self-supervising the consistency between the generated 3D meshes and final regressed 3D mesh. Due to the full 3D annotations are available, we only supervise the  $L_{pm}$ and $L_{3D}$.  

\subsection{D2S Projection\label{sec:D2S proj}}

As we introduced in Sec.\ref{sec:intro} and Sec. 3.3, available 3D datasets~\cite{Freihand2019,h36m} are always captured in a constrained environment with the fixed camera configurations. In this case, using in-the-wild 2D pose datasets~\cite{coco,mpii,aich} for training is a good choice for better generalization and adopted by many methods. To supervise the 3D pose of the estimated 3D human mesh with 2D pose, we need a camera model to make 3D-to-2D projection. In previous
works~\cite{hmr,gcmr,sun2019dsd-satn,monocualar,Freihand2019}, weak-perspective projection is widely adopted. Compared with the perspective projection that needs complex camera parameters (e.g. 3D translation $\boldsymbol{T}=(T_x, T_y, d)\in \mathbb{R}^{3}$, focal length $\boldsymbol{f}=(f_x, f_y)\in \mathbb{R}^{2}$, et al.), the weak-perspective one is much more simple. It only composes of two factors, scale $ s \in \mathbb{R}$ and 2D translation $ \boldsymbol{t}=(t_u, t_v)\in \mathbb{R}^{2}$ in image plane. The 2D projection $J^{p2D}_{i}=(u_i,v_i)$ of the $i$-th 3D keypoint $J_i=(x_i,y_i,z_i)$ (takes the body gravity center as the origin) can be simply derived in a weak-perspective manner using Eq.(\ref{eq:wpersp2}) and Eq.(\ref{eq:wpersp3}) while its perspective projection can be derived using Eq.(\ref{eq:persp1}) and Eq.(\ref{eq:persp2}).

\begin{figure}[t]
    \centering
    \includegraphics[width=0.95\columnwidth]{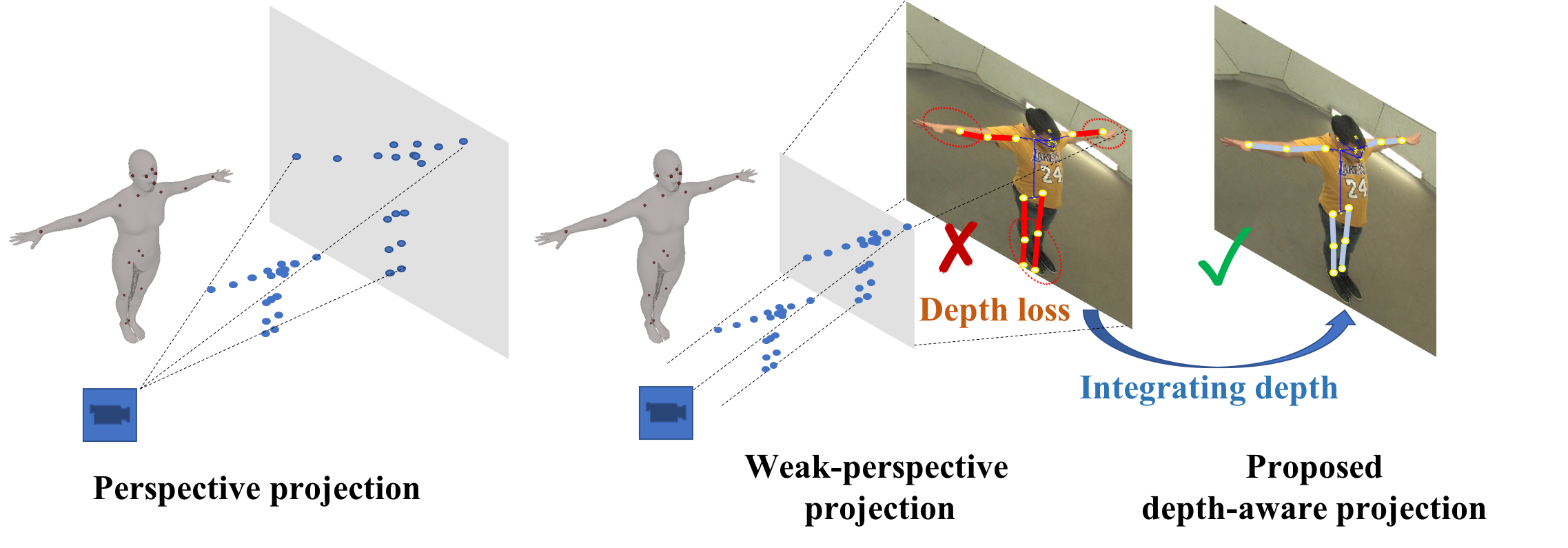}
    \caption{Effectiveness of the proposed D2S  projection. The widely used weak-perspective projection ignores the foreshortening effects in images. Through integrating depth differences among different keypoints, our proposed D2S projection successfully makes up the inherent depth loss.}
    \label{fig: D2S projection}
\end{figure}

We can notice that the weak-perspective projection shares the same scale $s$ among all keypoints, while the standard perspective projection provides an individual scale $\frac{f_x(x_i)}{d+z_i}$ for each keypoint. As shown in Fig.\ref{fig: D2S projection}, the weak-perspective projection ignores the difference induced by the foreshortening effect and results in 3D-to-2D misalignment. In this case, even if the 3D pose of the estimated human mesh is correct, the weak-perspective projection may lead to the wrong 2D pose supervision. Therefore, we develop a novel D2S projection to provide an individual projection scale for each keypoint w.r.t their depth difference. To unify the different representations, we assume that $s$ and $t_u$ are approximating the projection factor of a virtual body center whose $z_i=0$.
 In this way, the approximation relationship between the perspective and the weak-perspective projection can be represented as
\begin{equation}
\begin{aligned}
s = \frac{f_x}{d} \approx \frac{f_x}{d  + z_i}, 
\end{aligned}
\label{eq:comp1}
\end{equation}
\begin{equation}
\begin{aligned}
t_u = \frac{f_x t_x}{d} \approx \frac{f_x t_x}{d + z_i},
\end{aligned}
\label{eq:comp2}
\end{equation}

Based on this assumption, the difference ($\Delta s, \Delta t_u$) between the real and the approximate factor can be derived as 
\begin{equation}
\begin{aligned}
\Delta s = \frac{f_x}{d} - \frac{ f_x z_i}{d + z_i} = \frac{f_x}{d}\frac{d}{d + z_i}  , 
\end{aligned}
\label{eq:derive}
\end{equation}
\begin{equation}
\begin{aligned}
\Delta t_u = \frac{f_x t_x}{d} - \frac{ f_x t_x z_i}{d  + z_i} = \frac{f_x t_x}{d}\frac{d}{d + z_i} .
\end{aligned}
\label{eq:derive2}
\end{equation}

From Eq.(\ref{eq:comp1}) and  Eq.(\ref{eq:comp2}), we can know that, when $d$ is significantly larger than $z_i$, the difference between the perspective and the weak-perspective projection is small. Otherwise, the difference is obvious. In other words, when a person is close to the camera, the weak-perspective camera is prone to fail especially for poses with a large $z_i$. The difference is mainly brought by the depth $z_i$. Therefore, we design a D2S projection that converts the depth differences to projection scale variances among body parts. The goal is to make up the $\Delta s$ and  $\Delta t_u$, meanwhile, reduce the parameters that need to be estimated. 

Instead of  using a unified projection scale $\boldsymbol{s}$ and translation $t_u$ for all keypoints, the individual scale factor $\boldsymbol{s_i}$ for the $i$-th  keypoint is estimated to re-scale the output of weak-perspective projection. So the coordinate $u_i$ of the projected 2D keypoint $J^{2d}_i$ can be derived as
\begin{equation}
\begin{aligned}
u_i = s_i (s x_i + t_u),
\end{aligned}
\label{eq:dsp2}
\end{equation}
\begin{equation}
\begin{aligned}
s_i = \frac{d}{d + z_i},
\end{aligned}
\label{eq:scale}
\end{equation}
from which we can see that $d$ is a key coefficient converting the per-joint depth differences into the projection scale variation. In our framework, $d$ is estimated as a part of the camera parameters. Compared with the weak-perspective projection, we just need to estimate one more factor $d$. For better generalization on in-the-wild images, large-scale 2D pose datasets~\cite{mpii,aich} are used to learn the estimation of $d$.
The D2S projection works better in the scenarios whose depth difference between different parts is obvious and the field of view is large. For example, taking a selfie with a handheld camera. On the contrary, when people are far away from the camera, it degrades to the weak-perspective model. 

\section{Experiments}
We evaluate our proposed framework on three public large-scale benchmarks, CMU Panoptic Studio Dataset \cite{monocualar}, Human3.6M~\cite{h36m} body, and STB hand dataset~\cite{stb}. Following previous works~\cite{monocualar,smplx}, we also provide the visual results on expression estimation. Quantitative experiments favorably compare to the state-of the-art methods. Our qualitative results show accurate and robust recovery performance.

\subsection{Dataset\label{sec:dataset}}
We firstly introduce the three benchmarks for experiment evaluation. We also introduce several  datasets that we put together to establish the real distribution of whole-body parts for model training. 

\textbf{CMU Panoptic Studio Dataset}:
We use the ``Monocular Mocap'' part \cite{monocualar} that contains multi-view images (taken from 30 viewpoints) with 3D pose annotations. It contains $40$ subjects and $834K$ body images with corresponding 3D body pose annotations. Following~\cite{monocualar}, we only focus on the evaluation of 3D body pose in this paper. We report both mean per joint position error (MPJPE) and Procrustes Aligned MPJPE (PA-MPJPE) that is MPJPE after rigid alignment of the predicted pose with the ground truth.

\textbf{Human3.6M Dataset}: It is a large-scale human MoCap dataset ~\cite{h36m} with 3D pose annotations and has been widely used as the 3D pose benchmark. It contains videos of 7 actors performing 17 activities. Videos are captured in a controlled environment. For removing the redundancy, we down-sample all videos from 50fps to 10fps. We use the dataset to quantitatively evaluate the 3D body keypoint error of the recovered mesh and follow the same protocol as in~\cite{hmr} for evaluation. We evaluate both MPJPE and PA-MPJPE of the 3D body keypoints.

\textbf{STB Hand Dataset}: Stereo Hand Pose Tracking Benchmark (STB)~\cite{stb} is a 3D hand pose dataset, containing 18,000 images with corresponding depth images. The 3D hand annotations involve 21 keypoints locations. Following~\cite{hand_graph}, we split it into 15,000 samples for training and 3,000 samples for testing. Besides, for a fair comparison, we use Openpose~\cite{openpose} for bounding box detection. We measure the MPJPE for comparison.

\textbf{Datasets for STS and Model Generalization}:
For rationally sampling the SMPL-X model space, we use multiple 3D datasets to obtain the real whole-body parameters. 
1) 3D body poses are obtained from the AMASS dataset~\cite{amass}. It contains 300 subjects, more than 11000 motions (with corresponding SMPL 3D pose parameters), which is significantly richer than existing 3D pose datasets.
 2) 3D hand poses are obtained from the MANO~\cite{smplh}, Freihand~\cite{Freihand2019}, and AMASS~\cite{amass}. MANO and AMASS contain the 3D hand pose parameters. Freihand contains image data along with the 3D hand annotations. We split their training set into two sets for training (30000 samples) and validation (2560 samples) respectively.
3) Facial expression are extracted from D3DFACS 3D face mesh dataset~\cite{D3DFACS_dataset} and LS3D~\cite{bulat2017far} 2D facial dataset. Besides, we employ two in-the-wild 2D pose datasets, AI-CH~\cite{aich} and MPII~\cite{mpii},  to provide 2D pose supervision. Different from the 3D datasets captured in limited environments, they can greatly help the model generalize to various poses and camera configurations. 

\subsection{Implementation Details}
Considering that only a part of the data has 3D hand pose annotations, the STS is used one time in every 5 iterations. The Adam~\cite{adam} optimizer is adopted with betas=(0.9,0.999), momentum=0.9. The learning rate and batch size are set to 1e-4 and 16 respectively. The weights of loss items are set as $w_{pm}=20, w_{r}=0.5, w_{2D}=6, w_{3D}=60$. 
Training of the partial parameter retrieval branch consists of two phases. Firstly, we train the face and hand GCN-based regressor with the assistance of the FLAME  head model and MANO hand model respectively. Then, pre-trained models are loaded and the entire branch is trained by STS.

\subsection{Effect of D2S Projection.\label{dsp}}

In this section, we introduce the comprehensive ablation study of D2S projection on the CMU Panoptic Studio dataset. 
To validate the superior of the proposed D2S projection under an uncalibrated condition, we compare it with other camera models, which are the perspective projection (PP) and the weak-perspective projection (WPP) introduced in Sec.~\ref{sec:background}. For a fair comparison, the model is trained to estimate the camera parameters of these projection models based on the 2D pose only without using the ground truth parameters.  
Due to the complete camera parameters of perspective projection are too complex to predict, we simplify the parameters and estimate the $\frac{f_x + f_y}{2}, t^c$ in Eq.(\ref{eq:persp1}) and Eq.(\ref{eq:persp2}). For WPP, we just estimate the $s, t_u, t_v$ in Eq.(\ref{eq:wpersp2}) and Eq.(\ref{eq:wpersp3}). 

As shown in Tab.~\ref{PP_dataset}, \textit{PP} achieves comparative results with MTC, while greatly outperformed by the proposed D2S projection 8.6mm MPJPE and 4.1mm PA-MPJPE, corresponding to $13.9\%$ and $7.7\%$ error reduction. Especially, the training of the proposed D2S projection is much more stable than the \textit{PP}, which is prone to collapse.  \textit{WPP} is outperformed by 4.3mm MPJPE and 2.1mm PA-MPJPE, corresponding to $7.4\%$ and $5\%$ error reduction. We think the reason is that {PP} is too hard to estimate from a single monocular image, while the \textit{WPP} is so rough that may misguide the model. In contrast, the proposed D2S projection finds a good balance of estimation difficulty and projection accuracy for proper supervision. It demonstrates that the proposed D2S projection is more suitable for the un-calibrated condition in this task. 

\begin{figure}[t]
    \centering
    \includegraphics[width=0.98\columnwidth]{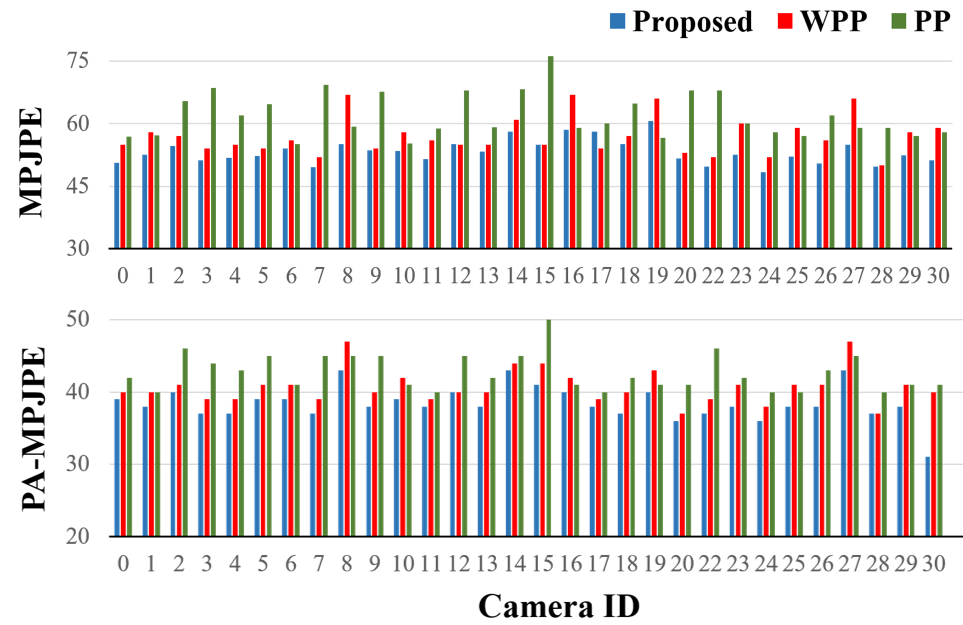}
    \caption{Comparisons between the perspective (\textit{PP}), the weak-perspective (\textit{WPP}) and the proposed D2S projection (\textit{Proposed}). We evaluate the MPJPE and PA-MPJPE (in millimetre) in different view points. }
    \label{fig:multiview}
\end{figure}

Moreover, since the Panoptic Studio Dataset contains multi-view images, we further evaluate the MPJPE and PA-MPJPE in different viewpoints. As shown in Fig.~\ref{fig:multiview}, the recovery error of the \textit{WPP} and \textit{PP} dramatically changes along with the viewpoint. The proposed D2S projection can alleviate this phenomenon to some degree, demonstrating that the proposed D2S projection is effective to make up the depth loss of weak-perspective projection.

\subsection{Comparisons to the State-of-the-arts}
\subsubsection{Evaluation on CMU Panoptic Studio Dataset.}
We compare our framework with MTC (monocular total capture)~\cite{monocualar} on all testing samples of CMU Panoptic Studio Dataset for evaluating 3D body pose accuracy. MTC deals with the same task as ours, while it makes the recovery of each human part independently by the combination of CNNs and optimization algorithms. As shown in the first two rows of Tab.~\ref{PP_dataset}, our proposed method outperforms MTC by 9.1mm in terms of MPJPE, corresponding to $14\%$ error reduction. It indicates the superiority of our single-step model-based framework over the optimization-based method. The results demonstrate the effectiveness of the proposed synthetic training strategy. Moreover, our proposed D2S projection further boosts the 3D body recovery performance.

\begin{table}[t]
    \setlength\tabcolsep{5pt}
    \centering
    \caption{\small {3D body pose evaluation on \textbf{Panoptic Studio Dataset}. The results of replacing the proposed D2S projection with the perspective projection and the weak-perspective projection are also presented. }}
\begin{tabular}{ccc}
        \hline
        Method & \small{MPJPE$(\downarrow)$} & \small{PA-MPJPE$(\downarrow)$} \\
        \hline
        MTC~\cite{monocualar} & {63.0} &{--} \\
        $\textbf{Proposed} $&  \textbf{53.2}   & \textbf{39.2}\\
        - Weak-perspective proj. & {57.5} & {41.3}\\
        - Perspective proj. & {61.8} & {43.3}\\
        \hline
    \end{tabular}%
    \label{PP_dataset}%
\end{table}

\begin{table}[t]
    \setlength\tabcolsep{8pt}
    \centering
    \caption{\small {Comparisons to state-of-the-art approaches on Human3.6M dataset.}}
    \begin{tabular}{ccc}
        \hline
       \textbf{Method} & \textbf{MPJPE} &\textbf{PA-MPJPE} \\
       \hline
        HMR~\cite{hmr} & {87.9} & {58.1} \\
        GCMR~\cite{gcmr} & {74.7} & {51.9}\\
        MTC~\cite{monocualar} & {58.3} & {-} \\
        SMPLify-X~\cite{smplx} & {-} &{75.9} \\
        $\textbf{Ours} $&   67.4   & {52.0} \\
        \hline
    \end{tabular}%
    \label{h36m}%
\end{table}%

\begin{table}[t]
    \setlength\tabcolsep{2pt}
    \centering
    \caption{\small {Evaluation on the \textbf{STB hand dataset}. FHGR denotes the face and hands GCN-based regressor in Fig.\ref{framework}. Ours means training the FHGR using the proposed STS simultaneously. $\star$ stands for using the ground truth 2D hand keypoints as input.}}
    \begin{tabular}{c|c|c|c|c}
        \hline
      \textbf{Method} &  GraphHand~\cite{hand_graph}  &  FHGR  &  \textbf{Ours} & $ \textbf{Ours}^\star $\\
       \hline
       \textbf{MPJPE $(\downarrow)$}  & 11.3 & 13.1  &  12.1 & 9.1\\
        \hline
    \end{tabular}%
    \label{hand_stb}%
\end{table}%

\begin{figure}[t]
    \centering
    \includegraphics[width=0.98\columnwidth]{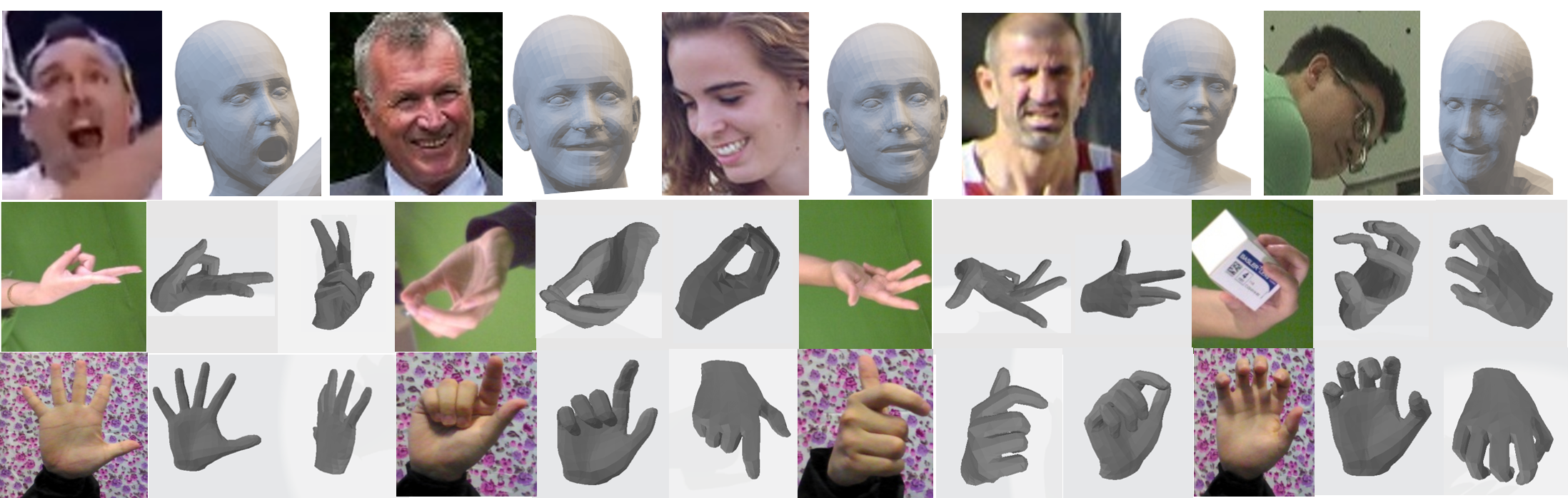}
    \caption{Qualitative results of face and hands. The top two rows are the recovered meshes on face data for both in indoor images from ~\cite{monocualar} and in-the-wild images. The third and fourth rows show the recovered hand meshes on the Freihand dataset~\cite{Freihand2019} and the STB dataset~\cite{stb}, respectively. For each input image, we show our recovered mesh in the front view and a random view.}
    \label{hand_result}
\end{figure}

\begin{figure}[t]
    \centering
    \includegraphics[width=0.98\columnwidth]{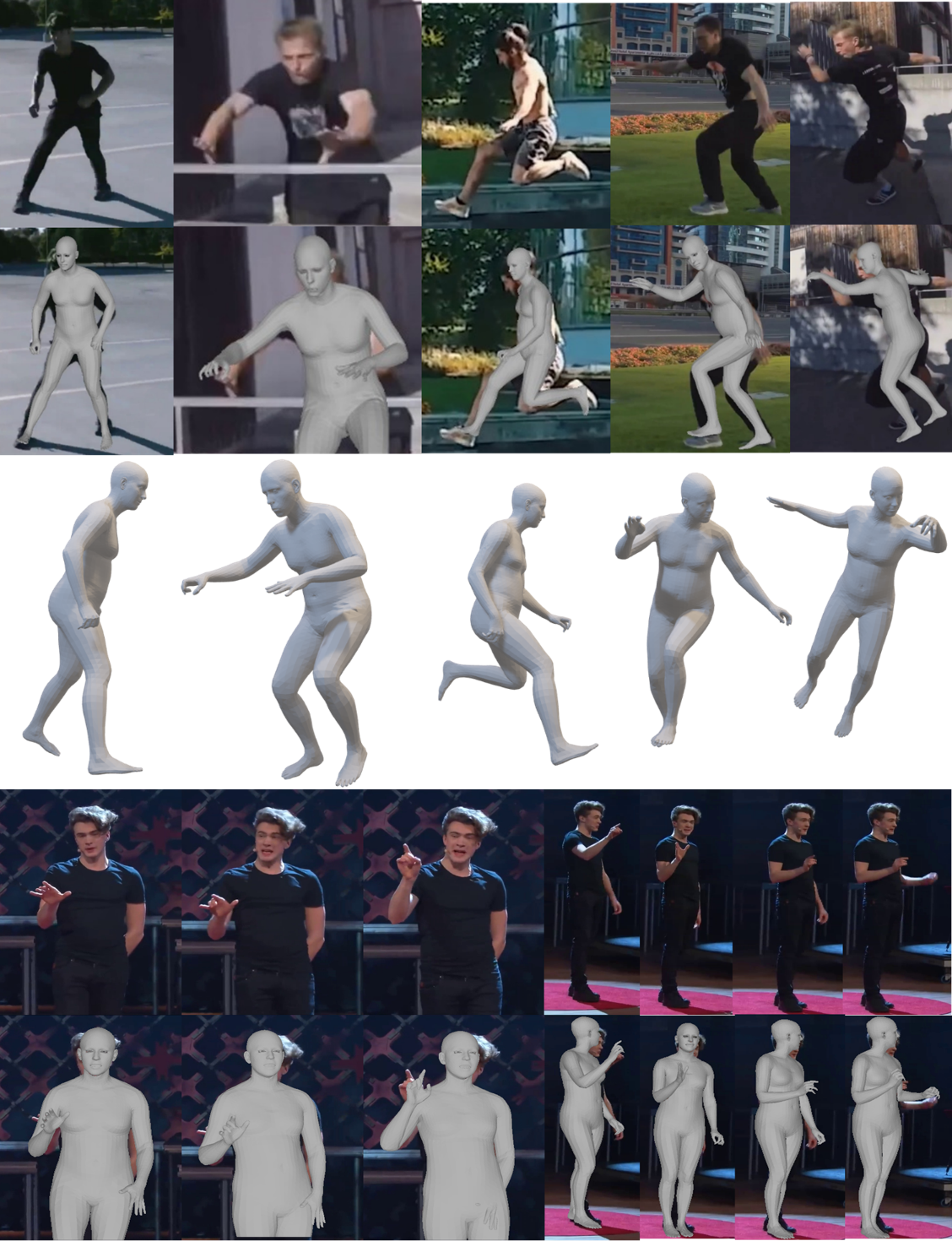}
    \caption{\textbf{Qualitative results on internet images where people are far from the camera.}}
    \label{fig:demo_internet_far}
\end{figure}

\subsubsection{Evaluation on Human3.6M Dataset.} 
As shown in Tab.\ref{h36m}, we compare our method with state-of-the-art methods on Human3.6M for 3D body keypoints estimation. Similar to us, the whole-body recovery method SMPLify-X~\cite{smplx} also fits the SMPL-X model, but it is an optimization-based algorithm. SMPLify-X is significantly outperformed by our single-step regressor method in terms of the PA-MPJPE. It demonstrates our performance. Furthermore, HMR~\cite{hmr} and GCMR~\cite{gcmr} are end-to-end methods for sole 3D body mesh recovery only. As shown in Tab.~\ref{h36m}, our method achieves competitive performance in terms of both MPJPE and PA-MPJPE. It demonstrates that our proposed method preserves the stable performance of the skeleton pose recovery, meanwhile, can accomplish the whole-body recovery. 
Besides, we also show some 3D pose estimation methods in the left part of Tab.~\ref{h36m}, which perform extremely well on this dataset, since they directly predict the 3D body keypoint other than 3D mesh. Here, we focus on the 3D mesh recovery methods and our method achieves comparable results. 

\subsubsection{Evaluation on STB Hand Dataset.}
We evaluate our method on the STB hand benchmark and present the MPJPE of hand keypoints.
As shown in Tab.~\ref{hand_stb}, our whole-body method can achieve comparable performance with GraphHand~\cite{hand_graph} which is the most recently proposed state-of-the-art method specially designed for hand recovery only. It demonstrates that our framework can well deal with the partial part recovery along with the whole-body setting.

To further evaluate the effectiveness of the partial part recovery, we conduct \textbf{ablation studies} on the partial parameter retrieval branch of our framework.
The face/hand GCN-based regressor for estimating partial parameters (see Fig.~\ref{framework}) is denoted as FHGR for simplification. In this dataset setting, we simply change the target of FHGR to estimate 3D hand pose from 2D hand landmarks. 
In Tab.~\ref{hand_stb}, the FHGR is solely trained on the STB dataset, while our method adds the synthetic training scheme (STS) to train the FHGR using synthetic data simultaneously. With the help of STS, FHGR moves towards GraphHand~\cite{hand_graph}, indicating that STS is vital for recovering the partial parameters when only unpaired data (body and hand/face) available.
Besides, the result using ground truth 2D pose in Tab.~\ref{hand_stb} shows that better 2D pose estimation leads to better performance.  

\begin{table}[t]
    \setlength\tabcolsep{2pt}
    \centering
    \caption{\small {Comparisons to state-of-the-art methods on computational efficiency. }}
    \begin{tabular}{cccc}
        \hline
        Method & SMPLify-X~\cite{smplx} & MTC~\cite{monocualar}&  \textbf{Ours} \\
        \hline
        Avg. runtime(s) & 12 & 82 & \textbf{0.09}\\
        \hline
    \end{tabular}
    \label{runtime}
\end{table}

\subsection	{Efficiency analysis}
In Tab.\ref{runtime}, we compare the computational efficiency with the most related works SMPLify-X~\cite{smplx} and MTC~\cite{monocualar}. 
For a fair comparison, we directly use the released official code and test on the same dataset~\cite{FECdataset} on the same device (Tesla P40 GPU). Considering that both SMPLify-X and our method employ the OpenPose to estimate the whole-body 2D keypoints, we only compare the time consuming of predicting the whole-body 3D mesh from a single 2D image and the corresponding OpenPose output. SMPLify-X takes 82.72 seconds per image, while the proposed single-step framework only consumes 0.09 seconds per image. The time consumption is mainly costed by the 5-stage iterative optimization strategy in their methods, which demonstrate our motivation.
Based on their official code, MTC costs about 12 seconds per image. The single-step design enables us to enjoy parallel computation, which greatly accelerates the calculation. In contrast, the proposed single-step method can usually be dozens or hundreds of times faster than the existing optimization-based methods.

\subsection{Qualitative Evaluation}
We present qualitative results of our method on images for the whole-body mesh recovery and the face/hand partial recovery.
 To test the effectiveness and the stability of the proposed method (STS and D2S projection), we evaluate our model on internet images. The results of MTC~\cite{monocualar}, SMPLify-X~\cite{smplx} and the proposed method are presented in Fig.\ref{fig:demo1}. 
MTC and SMPLify-X can indeed align the human body in 2D images better, but their performance in depth and shape estimation is difficult to satisfy. 
Especially, when some body parts are invisible (such as the invisible feet at the first raw of Fig.\ref{fig:demo1}), the algorithm is prone to fail. The estimated body shape is weird and not coordinated. In contrast, the estimated body pose, shape, and expression of our method are much better. Benefited from the D2S projection, our method shows some prior in perceiving depth information.
Considering most 3D datasets are captured in a close range, we deliberately select some videos where people are far from the camera. The results are presented in Fig.\ref{fig:demo_internet_far}. It demonstrates that the proposed D2S projection can handle a wide range of human-to-camera distance.
Besides, qualitative results of face and hands are shown in Fig.~\ref{hand_result}. The top row presents the recovered meshes on face data for both in indoor images from~\cite{monocualar} and in-the-wild images. The second and third rows show the recovered hand meshes on the Freihand dataset~\cite{Freihand2019} and the STB dataset~\cite{stb}, respectively.  These results demonstrate the effectiveness of the proposed method.

\subsection{Discussion}

The proposed method successfully reduces the network's dependence on data and improves computational efficiency. However, there are still some limitations. 

\textbf{Reliance on 2D pose estimation.} The performance relies on the quality of the 2D pose estimation. A stronger 2D pose estimation model will lead to better performance. Also, the representation learning of the body shape needs to be further enhanced. Besides, self-occlusion, such as hands overlapping, also affects the performance. We will investigate those problems in our further work.

\textbf{About face shape. } Besides, due to the limitation of the SMPL-X model, we only estimate the facial expression instead of the 3D face mesh. Because the face shape of SMPL-X is changed along with the body shape and cannot be controlled individually, which makes it hard to estimate the precise facial mesh. Therefore, we show the visualization results other than evaluating the facial keypoints. We will dig into developing a whole-body statistical model whose partial mesh could be preciously controlled.

\section{Conclusion}
In this paper, we present a synthetic training framework for learning a single-step regression model that jointly recovers the mesh of multiple 3D human body parts from a single whole-body image using unpaired data. Compared with the previous optimization-based method, the proposed method shows an obvious advantage in computational efficiency.
To deal with the large scale difference among different body parts, the network architecture is designed in a disentangled manner. It facilitates the individual learning of different body properties.
Additionally, the depth loss of weak-perspective projection is also investigated and a D2S projection is proposed to learn the projection distortion. With the D2S projection, we exploit the 2D pose supervision for generalization and rich representation. Extensive experiments with ablation analysis show that our proposed method achieves superior performance on relevant benchmarks.

\bibliographystyle{IEEEtran}
\bibliography{egbib.bib}

\begin{thebibliography}{10}
\providecommand{\url}[1]{#1}
\csname url@samestyle\endcsname
\providecommand{\newblock}{\relax}
\providecommand{\bibinfo}[2]{#2}
\providecommand{\BIBentrySTDinterwordspacing}{\spaceskip=0pt\relax}
\providecommand{\BIBentryALTinterwordstretchfactor}{4}
\providecommand{\BIBentryALTinterwordspacing}{\spaceskip=\fontdimen2\font plus
\BIBentryALTinterwordstretchfactor\fontdimen3\font minus
  \fontdimen4\font\relax}
\providecommand{\BIBforeignlanguage}[2]{{%
\expandafter\ifx\csname l@#1\endcsname\relax
\typeout{** WARNING: IEEEtran.bst: No hyphenation pattern has been}%
\typeout{** loaded for the language `#1'. Using the pattern for}%
\typeout{** the default language instead.}%
\else
\language=\csname l@#1\endcsname
\fi
#2}}
\providecommand{\BIBdecl}{\relax}
\BIBdecl

\bibitem{de2008performance}
E.~De~Aguiar, C.~Stoll, C.~Theobalt, N.~Ahmed, H.-P. Seidel, and S.~Thrun,
  ``Performance capture from sparse multi-view video,'' in \emph{ACM SIGGRAPH},
  2008, pp. 1--10.

\bibitem{vlasic2008articulated}
D.~Vlasic, I.~Baran, W.~Matusik, and J.~Popovi{\'c}, ``Articulated mesh
  animation from multi-view silhouettes,'' in \emph{ACM SIGGRAPH}, 2008.

\bibitem{keep}
F.~Bogo, A.~Kanazawa, C.~Lassner, P.~Gehler, J.~Romero, and M.~J. Black, ``Keep
  it smpl: Automatic estimation of 3{D} human pose and shape from a single
  image,'' in \emph{ECCV}, 2016.

\bibitem{hmr}
A.~Kanazawa, M.~J. Black, D.~W. Jacobs, and J.~Malik, ``End-to-end recovery of
  human shape and pose,'' in \emph{IEEE CVPR}, 2018.

\bibitem{gcmr}
N.~Kolotouros, G.~Pavlakos, and K.~Daniilidis, ``Convolutional mesh regression
  for single-image human shape reconstruction,'' in \emph{IEEE CVPR}, 2019.

\bibitem{humanshape}
G.~Pavlakos, L.~Zhu, X.~Zhou, and K.~Daniilidis, ``Learning to estimate 3{D}
  human pose and shape from a single color image,'' in \emph{IEEE CVPR}, 2018.

\bibitem{sun2019dsd-satn}
Y.~Sun, Y.~Ye, W.~Liu, W.~Gao, Y.~Fu, and T.~Mei, ``Human mesh recovery from
  monocular images via a skeleton-disentangled representation,'' in \emph{IEEE
  ICCV}, 2019.

\bibitem{flame}
T.~Li, T.~Bolkart, M.~J. Black, H.~Li, and J.~Romero, ``Learning a model of
  facial shape and expression from {4D} scans,'' \emph{ACM SIGGRAPH Asia},
  2017.

\bibitem{ringnet}
S.~Sanyal, T.~Bolkart, H.~Feng, and M.~J. Black, ``Learning to regress 3d face
  shape and expression from an image without 3d supervision,'' in \emph{IEEE
  CVPR}, 2019.

\bibitem{zollhofer2018state}
M.~Zollh{\"o}fer, J.~Thies, P.~Garrido, D.~Bradley, T.~Beeler, P.~P{\'e}rez,
  M.~Stamminger, M.~Nie{\ss}ner, and C.~Theobalt, ``State of the art on
  monocular 3d face reconstruction, tracking, and applications,'' in
  \emph{Computer Graphics Forum}, 2018.

\bibitem{hand_graph}
L.~Ge, Z.~Ren, Y.~Li, Z.~Xue, Y.~Wang, J.~Cai, and J.~Yuan, ``3d hand shape and
  pose estimation from a single rgb image,'' in \emph{IEEE CVPR}, 2019.

\bibitem{de2011model}
M.~de~La~Gorce, D.~J. Fleet, and N.~Paragios, ``Model-based 3d hand pose
  estimation from monocular video,'' \emph{IEEE Transactions on Pattern
  Analysis and Machine Intelligence}, 2011.

\bibitem{smplh}
J.~Romero, D.~Tzionas, and M.~J. Black, ``Embodied hands: Modeling and
  capturing hands and bodies together,'' \emph{ACM Transactions on Graphics},
  2017.

\bibitem{monocualar}
D.~Xiang, H.~Joo, and Y.~Sheikh, ``Monocular total capture: Posing face, body,
  and hands in the wild,'' in \emph{IEEE CVPR}, 2019.

\bibitem{smplx}
G.~Pavlakos, V.~Choutas, N.~Ghorbani, T.~Bolkart, A.~A. Osman, D.~Tzionas, and
  M.~J. Black, ``Expressive body capture: 3d hands, face, and body from a
  single image,'' in \emph{IEEE CVPR}, 2019.

\bibitem{totalcapture}
H.~Joo, T.~Simon, and Y.~Sheikh, ``Total capture: A 3d deformation model for
  tracking faces, hands, and bodies,'' in \emph{IEEE CVPR}, 2018.

\bibitem{stb}
J.~Zhang, J.~Jiao, M.~Chen, L.~Qu, X.~Xu, and Q.~Yang, ``3d hand pose tracking
  and estimation using stereo matching,'' \emph{arXiv:1610.07214}, 2016.

\bibitem{Freihand2019}
Z.~Christian, C.~Duygu, Y.~Jimei, R.~Bryan, A.~Max, and B.~Thomas, ``Freihand:
  A dataset for markerless capture of hand pose and shape from single rgb
  images,'' in \emph{IEEE ICCV}, 2019.

\bibitem{FECdataset}
R.~Vemulapalli and A.~Agarwala, ``A compact embedding for facial expression
  similarity,'' in \emph{IEEE CVPR}, 2019, pp. 5683--5692.

\bibitem{D3DFACS_dataset}
D.~Cosker, E.~Krumhuber, and A.~Hilton, ``A facs valid 3d dynamic action unit
  database with applications to 3d dynamic morphable facial modeling,'' in
  \emph{IEEE ICCV}, 2011.

\bibitem{h36m}
C.~Ionescu, D.~Papava, V.~Olaru, and C.~Sminchisescu, ``Human3.6{M}: Large
  scale datasets and predictive methods for 3{D} human sensing in natural
  environments,'' \emph{IEEE Transactions on Pattern Analysis and Machine
  Intelligence}, 2014.

\bibitem{openpose}
Z.~Cao, T.~Simon, S.-E. Wei, and Y.~Sheikh, ``Realtime multi-person 2d pose
  estimation using part affinity fields,'' in \emph{IEEE CVPR}, 2017.

\bibitem{whole-2d-openpose}
G.~Hidalgo, Y.~Raaj, H.~Idrees, D.~Xiang, H.~Joo, T.~Simon, and Y.~Sheikh,
  ``Single-network whole-body pose estimation,'' in \emph{IEEE ICCV}, 2019.

\bibitem{TIP_2DPEnie2018hierarchical}
X.~C. Nie, J.~S. Feng, J.~L. Xing, S.~T. Xiao, and S.~C. Yan, ``Hierarchical
  contextual refinement networks for human pose estimation,'' \emph{IEEE
  Transactions on Image Processing}, pp. 924--936, 2018.

\bibitem{TIP_2DPEluo2018multi}
Y.~M. Luo, Z.~T. Xu, P.~Z. Liu, Y.~Z. Du, and J.~M. Guo, ``Multi-person pose
  estimation via multi-layer fractal network and joints kinship pattern,''
  \emph{IEEE Transactions on Image Processing}, pp. 142--155, 2018.

\bibitem{TIP_2DPEfu2016orgm}
L.~R. Fu, J.~G. Zhang, and K.~Q. Huang, ``Orgm: occlusion relational graphical
  model for human pose estimation,'' \emph{IEEE Transactions on Image
  Processing}, pp. 927--941, 2016.

\bibitem{chen20173d}
C.-H. Chen and D.~Ramanan, ``3d human pose estimation= 2d pose estimation+
  matching,'' in \emph{IEEE CVPR}, 2017.

\bibitem{fang2018learning}
H.-S. Fang, Y.~Xu, W.~Wang, X.~Liu, and S.-C. Zhu, ``Learning pose grammar to
  encode human body configuration for 3d pose estimation,'' in \emph{AAAI},
  2018.

\bibitem{nie2017monocular}
B.~X. Nie, P.~Wei, and S.-C. Zhu, ``Monocular 3d human pose estimation by
  predicting depth on joints,'' in \emph{IEEE ICCV}, 2017.

\bibitem{soft-argmax}
D.~C. Luvizon, D.~Picard, and H.~Tabia, ``2d/3d pose estimation and action
  recognition using multitask deep learning,'' in \emph{IEEE CVPR}, 2018.

\bibitem{sun2017compositional}
X.~Sun, J.~Shang, S.~Liang, and Y.~Wei, ``Compositional human pose
  regression,'' in \emph{IEEE ICCV}, 2017.

\bibitem{zhou2017towards}
X.~Zhou, Q.~Huang, X.~Sun, X.~Xue, and Y.~Wei, ``Towards 3d human pose
  estimation in the wild: a weakly-supervised approach,'' in \emph{IEEE ICCV},
  2017.

\bibitem{TIP_3DPEzheng2020joint}
X.~T. Zheng, X.~M. Chen, and X.~Q. Lu, ``A joint relationship aware neural
  network for single-image 3d human pose estimation,'' \emph{IEEE Transactions
  on Image Processing}, pp. 4747--4758, 2020.

\bibitem{smpl}
M.~Loper, N.~Mahmood, J.~Romero, G.~Pons-Moll, and M.~J. Black, ``{SMPL}: A
  skinned multi-person linear model,'' \emph{ACM Transactions on Graphics},
  2015.

\bibitem{yoshiyasu2018skeleton}
Y.~Yoshiyasu, R.~Sagawa, K.~Ayusawa, and A.~Murai, ``Skeleton transformer
  networks: 3d human pose and skinned mesh from single rgb image,''
  \emph{arXiv:1812.11328}, 2018.

\bibitem{monocap}
X.~Zhou, M.~Zhu, G.~Pavlakos, S.~Leonardos, K.~G. Derpanis, and K.~Daniilidis,
  ``Monocap: Monocular human motion capture using a cnn coupled with a
  geometric prior,'' \emph{IEEE Transactions on Pattern Analysis and Machine
  Intelligence}, 2018.

\bibitem{monoperfcap}
W.~Xu, A.~Chatterjee, M.~Zollh{\"o}fer, H.~Rhodin, D.~Mehta, H.-P. Seidel, and
  C.~Theobalt, ``Monoperfcap: Human performance capture from monocular video,''
  \emph{ACM Transactions on Graphics}, 2018.

\bibitem{Xu_2019_ICCV}
Y.-L. Xu, S.-C. Zhu, and T.~Tung, ``Denserac: Joint 3d pose and shape
  estimation by dense render-and-compare,'' in \emph{IEEE ICCV}, 2019.

\bibitem{pavlakos2019texturepose}
G.~Pavlakos, N.~Kolotouros, and K.~Daniilidis, ``Texturepose: Supervising human
  mesh estimation with texture consistency,'' in \emph{IEEE ICCV}, 2019.

\bibitem{Kundu_Appearance_Consensus}
K.~Jogendra, Nath, R.~Mugalodi, J.~Varun, V.~Rahul, Mysore, and B.~R.,
  Venkatesh, ``Appearance consensus driven self-supervised human mesh
  recovery,'' in \emph{ECCV}, 2020.

\bibitem{Choi_Pose2Mesh}
C.~Hongsuk, M.~Gyeongsik, and L.~Kyoung, Mu, ``Pose2mesh: Graph convolutional
  network for 3d human pose and mesh recovery from a 2d human pose,'' in
  \emph{ECCV}, 2020.

\bibitem{Kulon_2020_CVPR}
D.~Kulon, R.~A. Guler, I.~Kokkinos, M.~M. Bronstein, and S.~Zafeiriou,
  ``Weakly-supervised mesh-convolutional hand reconstruction in the wild,'' in
  \emph{IEEE CVPR}, 2020.

\bibitem{zhou2020monocular}
Y.~Zhou, M.~Habermann, W.~Xu, I.~Habibie, C.~Theobalt, and F.~Xu, ``Monocular
  real-time hand shape and motion capture using multi-modal data,'' in
  \emph{IEEE CVPR}, 2020.

\bibitem{resnet}
K.~He, X.~Zhang, S.~Ren, and J.~Sun, ``Deep residual learning for image
  recognition,'' in \emph{IEEE CVPR}, 2016.

\bibitem{zhao2019semantic}
L.~Zhao, X.~Peng, Y.~Tian, M.~Kapadia, and D.~N. Metaxas, ``Semantic graph
  convolutional networks for 3d human pose regression,'' in \emph{IEEE CVPR},
  2019.

\bibitem{bulat2017far}
A.~Bulat and G.~Tzimiropoulos, ``How far are we from solving the 2d \& 3d face
  alignment problem?'' in \emph{IEEE ICCV}, 2017.

\bibitem{amass}
N.~Mahmood, N.~Ghorbani, N.~F. Troje, G.~Pons-Moll, and M.~J. Black, ``{AMASS}:
  Archive of motion capture as surface shapes,'' in \emph{IEEE ICCV}, 2019.

\bibitem{coco}
T.-Y. Lin, M.~Maire, S.~Belongie, J.~Hays, P.~Perona, D.~Ramanan,
  P.~Doll{\'a}r, and C.~L. Zitnick, ``Microsoft coco: Common objects in
  context,'' in \emph{ECCV}, 2014.

\bibitem{mpii}
M.~Andriluka, L.~Pishchulin, P.~Gehler, and B.~Schiele, ``2d human pose
  estimation: New benchmark and state of the art analysis,'' in \emph{IEEE
  CVPR}, 2014.

\bibitem{aich}
J.~Wu, H.~Zheng, B.~Zhao, Y.~Li, B.~Yan, R.~Liang, W.~Wang, S.~Zhou, G.~Lin,
  Y.~Fu \emph{et~al.}, ``Ai challenger: A large-scale dataset for going deeper
  in image understanding,'' \emph{arXiv:1711.06475}, 2017.

\bibitem{adam}
D.~P. Kingma and J.~Ba, ``Adam: A method for stochastic optimization,'' in
  \emph{ICLR}, 2015.

\end{thebibliography}

\end{document}